\documentclass[sigconf]{acmart}
\usepackage{todonotes}
\usepackage{multirow}
\usepackage{algorithm}
\usepackage{algorithmic}
\usepackage{latexsym}
\usepackage{graphicx}
\usepackage{soul} 
\usepackage{color, xcolor} 
\usepackage{framed}
\usepackage{makecell}
\usepackage{pifont}
\usepackage{booktabs}
\usepackage{subcaption}
\usepackage[most]{tcolorbox}

\AtBeginDocument{%
  \providecommand\BibTeX{{%
    \normalfont B\kern-0.5em{\scshape i\kern-0.25em b}\kern-0.8em\TeX}}}

\setcopyright{acmcopyright}
\copyrightyear{2018}
\acmYear{2024}
\acmDOI{XXXXXXX.XXXXXXX}

\acmConference[Conference acronym 'XX]{Make sure to enter the correct
  conference title from your rights confirmation emai}{June 03--05,
  2018}{Woodstock, NY}
\acmPrice{15.00}
\acmISBN{978-1-4503-XXXX-X/18/06}

\begin{document}

\title{Can Large Language Models Detect Misinformation in Scientific News Reporting?}


\author{Yupeng Cao}
\email{ycao33@stevens.edu}
\affiliation{%
  \institution{Stevens Institute of Technology}
  \city{Hoboken}
  \state{NJ}
  \country{USA}}

\author{Aishwarya Muralidharan Nair}
\email{anair9@stevens.edu}
\affiliation{%
  \institution{Stevens Institute of Technology}
  \city{Hoboken}
  \state{NJ}
  \country{USA}}

\author{Nastaran Jamalipour Soofi}
\email{njamalipour@stevens.edu}
\affiliation{%
  \institution{Stevens Institute of Technology}
  \city{Hoboken}
  \state{NJ}
  \country{USA}}

\author{Elyon Eyimife}
\email{eeyimife@stevens.edu}
\affiliation{%
  \institution{Stevens Institute of Technology}
  \city{Hoboken}
  \state{NJ}
  \country{USA}}

\author{K.P. Subbalakshmi}
\email{ksubbala@stevens.edu}
\affiliation{%
  \institution{Stevens Institute of Technology}
  \city{Hoboken}
  \state{NJ}
  \country{USA}}

\renewcommand{\shortauthors}{Cao et al.}

\begin{abstract}
\label{sec:abstract}
Automatic detection of misinformation in the scientific domain is challenging because of the distinct styles of writing in scientific publications vs reporting. 
This problem is exacerbated by the prevalence of large language model generated misinformation. In this paper, we address the problem of automatic detection of 
misinformation in a more realistic scenario where there is no prior knowledge of the origin (LLM or human written) of the text, and explicit claims may not be available. We first introduce a novel labeled dataset, \textbf{\textsc{CoSMis}}  {\textsc{(SciNews)}}, comprising of 2,400 scientific news stories sourced from both reliable and unreliable outlets, paired with relevant abstracts from the CORD-19 database. Our dataset uniquely includes both human-written and LLM-generated news articles. We propose a set of dimensions of scientific validity (DoV) along which to evaluate the articles for misinformation. These are then incorporated into the prompt structures for the LLMs. We propose three LLM pipelines to compare scientific news to relevant research papers and classify for misinformation. The three pipelines represent different levels of intermediate processing steps on the raw scientific news articles and research papers. We apply various prompt engineering strategies: zero-shot, few-shot, and DoV-guided Chain-of-Thought prompting, to these architectures and evaluate them using GPT-3.5, GPT-4, Llama2-7B/13B/70B and Llama3-8B.\footnote{This work has been accepted by AAAI-25 Workshop on Preventing and Detecting LLM Misinformation (\href{https://openreview.net/pdf?id=ZMewgcXcj1}{Link}).}\footnote{We release the dataset at: \url{https://github.com/InfintyLab/CoSMis-SciNews-}}
\end{abstract}





\maketitle

\section{Introduction} 
\label{sec:intro}
Scientific information is communicated to the non-expert audience via popular press (news articles) and online platforms like blogs, social media posts, etc. 
Studies have shown that news with 
scientific-sounding content is trusted more than other types~\cite{ShaEtal20}. 
Therefore, any misinformation in the scientific domain can cause significant public risk as was evidenced during the recent COVID-19 pandemic~\cite{MheEtal20}.
Other examples include
the emergence of vaccination hesitancy~\cite{RazEtal21, BaiEtal21}, eroded trust 
in health institutions~\cite{RodEtal21}, and the amplification of public fear and anxiety~\cite{NelEtal20, VerEtal22}. 
Therefore, it is imperative to identify 
misinformation in scientific news reporting. 

Several websites are maintained by science reporters (Health News Review\footnote{\url{https://www.healthnewsreview.org/}})
and scientists (Science Feedback \footnote{\url{https://science.feedback.org/}}) to track scientific misinformation in the media.
Although such manual debunking is important, the 
sheer volume of scientific news can make this task 
unscalable. Natural language processing (NLP) based approaches
have consequently started to emerge to deal with 
this problem. These methods typically involve 
language analysis, like detecting exaggeration~\cite{WriEtal21}
and certainty~\cite{PeiEtal21} and fact-checking~\cite{ZhiEtal22} and claim verification~\cite{RonEtal20}. Several claim verification datasets have also been developed for this problem~\cite{ThoEtal18, WaEtal22}
and a method for modeling information change from scientific article to scientific reporting has also been proposed~\cite{DusEtal22}.

While these works have laid the foundation 
to address this problem, the area is still in
its nascence. Several challenges remain unaddressed: 1) there are no existing taxonomies
to define dimensions of the scientific validity of 
scientific news articles that can be used in 
automated methods to detect misinformation in 
scientific news; 2) all existing datasets for 
scientific fact-checking relies on
explicit claim generation from the news articles
before it can be compared to the scientific articles
for misinformation detection. This can be a cumbersome process in real-life scenarios where it
would be potentially necessary for expert human 
involvement to first generate claims from the 
scientific article and 3) there is no generalized architecture that can detect scientific misinformation without an explicit claim generation
step.


In response to the aforementioned limitations, 
and because of the quantum leap in performance 
improvement offered by large language models (LLMs) in downstream NLP tasks, we 
formulate the following research questions:
\begin{itemize}
    \item RQ1: Can LLMs be used to define a general architecture to detect misinformation in scientific news reporting in simulated real-life scenarios without the need for explicit claims?
    \item RQ2: Is it feasible to define dimensions along which the scientific validity of the news article can be measured and aid in the creation of effective prompts for these architectures?
    \item RQ3: Do these architectures possess the capability to provide explanations for their decision-making processes?
\end{itemize}

To answer the above questions, we first 
create a novel \underline{CO}VID related \underline{S}cientific \underline{Mis}information (\textbf{\textbf{\textsc{CoSMis}, or SciNews}}\footnote{\url{https://github.com/InfintyLab/CoSMis-SciNews-}}) dataset, comprised of 
scientific news and related scientific articles.
Given the rising trend in LLM-generated content 
in both legitimate reporting and misinformation,
this dataset contains an equal number of LLM-generated and human-written articles.
The dataset construction pipeline is flexible enough to allow
continuous updates with emerging news articles and scientific articles.

We then propose three architectures using LLMs to 
automatically detect false representations of 
scientific findings in the popular press without 
explicit claim generation. The first architecture, SERIf, uses three modules: \textbf{S}ummarization, \textbf{E}vidence \textbf{R}etrieval, and \textbf{I}n\textbf{f}erence to classify the news article as fake or true; the second architecture, SIf, bypasses the explicit evidence retrieval module while keeping the other two, and the third, direct-to-inference architecture, D2I, adopts a direct-to-inference approach, dispensing with both summarization and explicit evidence retrieval. For each of these architectures, we employ several prompt engineering strategies including zero-shot, few-shot, and chain-of-thought prompting. We test these architectures using several state-of-the-art LLMs, including GPT (3.5\&4), Llama2 (7B,13B\&70B) and Llama3 (8B). 

This work makes the following contributions: 
1) introducing \textbf{\textsc{CoSMis} (SciNews)}, a unique dataset designed for detecting scientific misinformation, which includes human-authored articles and LLM-generated texts to mirror real-world challenges.
2) proposing three LLM pipelines to detect scientific misinformation ``in the wild" using scientific articles as grounding evidence material. 
3) proposing Dimensions of Validity (DoV) guided chain-of-thought prompting
4) testing the proposed pipelines on the architectures on the \textbf{\textsc{CoSMis}} dataset and demonstrating that LLMs are able to detect scientific misinformation without needing a training phase and testing phase and 5) demonstrating that the DoV prompting can be used to derive explanations for the LLM's decision.


\section{Related Work}
\label{sec:related}

As mentioned earlier, scientific misinformation
detection is still in its nascence and while 
related to misinformation detection in general, 
it is a harder problem since the language 
characteristics of the scientific communication is
different from the formal format of scientific publications. 
The problem of scientific misinformation
is related to two other concepts in NLP, including: 
1) fact-checking (claim verification) 2) scientific language analysis.

However, none of these approaches can singly 
capture the complexity of scientific misinformation and so far, there has not been any attempt to systematically capture the ways in which scientific misinformation can occur and then to use that to detect scientific misinformation.
In this work, we first define some dimensions
of scientific validity and then harness the power 
of LLMs to design general architectures to 
analyze scientific news for misinformation.

\subsection{Fact-Checking}
Automatic fact-checking, which assesses the truthfulness of claims made in the text~\cite{VlaRie14, GuoEtal22}, has been extensively studied across various domains, including common knowledge verification~\cite{ThoEtal18}, political topics~\cite{Wan17}, COVID-19~\cite{SaChMu21}, E-commercial~\cite{ZhaEtal20E}, biology~\cite{WenEtal22b}. When thinking about scientific misrepresentation in popular media,
it is natural to think of the veracity of scientific 
findings. It is therefore possible to cast the scientific misinformation as a fact-checking problem or claim verification problem.
Several researchers have taken this approach to defect misinformation in the scientific domain~\cite{VlaMat23, WaEtal20, ThoEtal20, WadKyl21, WaEtal22, WriEtal22, WaEtal23}. These works typically construct claims from the existing scientific literature by manually reformulated scientific findings and then the constructed claims are verified by utilizing pre-existing knowledge resources. However, most of these works rely on human resources to identify and extract appropriate claims for verification. Furthermore, these artificially constructed claims may not accurately represent the complexity and nuance of claims encountered in real-world scenarios.


\subsection{Scientific Language Analysis}
Scholarly document processing has garnered considerable attention in recent years, reflecting a growing interest in the analysis and interpretation of scientific literature~\cite{ChaEtal20overview}. Of particular relevance to our research are tasks that track the change of scientific information from published literature to social press. This includes investigating writing strategies employed in science communication ~\cite{TanLee14, AugEtal20}, detecting changes in certainty~\cite{PeiEtal21} and exaggeration detection~\cite{WriIsa21}. Furthermore, the automatic detection of semantic similarities between scientific texts and their paraphrases represents an alternative approach for analyzing scientific content~\cite{KylEtak19, JakEtal23}. 
However, rather than use the typical metrics for measuring semantic similarity, we propose to use the 
inherent knowledge in pre-trained 
LLMs for this task.

\subsection{Large Language Model in Misinformation}
\label{ssec:llm-mis}
LLMs have consistently demonstrated the ability to generate text on par with human authors ~\cite{ZhaEtal23, YanEtal23}. This has led to
their widespread use by professionals in generating legitimate real news stories. 
Unfortunately, they have also been used to generate misinformation~\cite{ZhoEtal23, CheShu23, Buc21Etal} and often at a much larger scale than is humanly possible. 
While falsehoods crafted by LLMs prove challenging 
for humans to detect, compared to human-generated ones~\cite{CheShu23}, 
several studies have illustrated the feasibility of 
identifying LLM-generated text using ~\cite{TanChuHu23, Buc21Etal}. 
Motivated by these studies, we include a balanced 
set of LLM-generated scientific articles, both 
fake and true, in the SciNews (\textsc{CoSMis})dataset.

\label{sec:data}
\section{CosMIs Dataset Construction}
\begin{figure*}[htbp]
\centering
\includegraphics[width=0.9\textwidth]{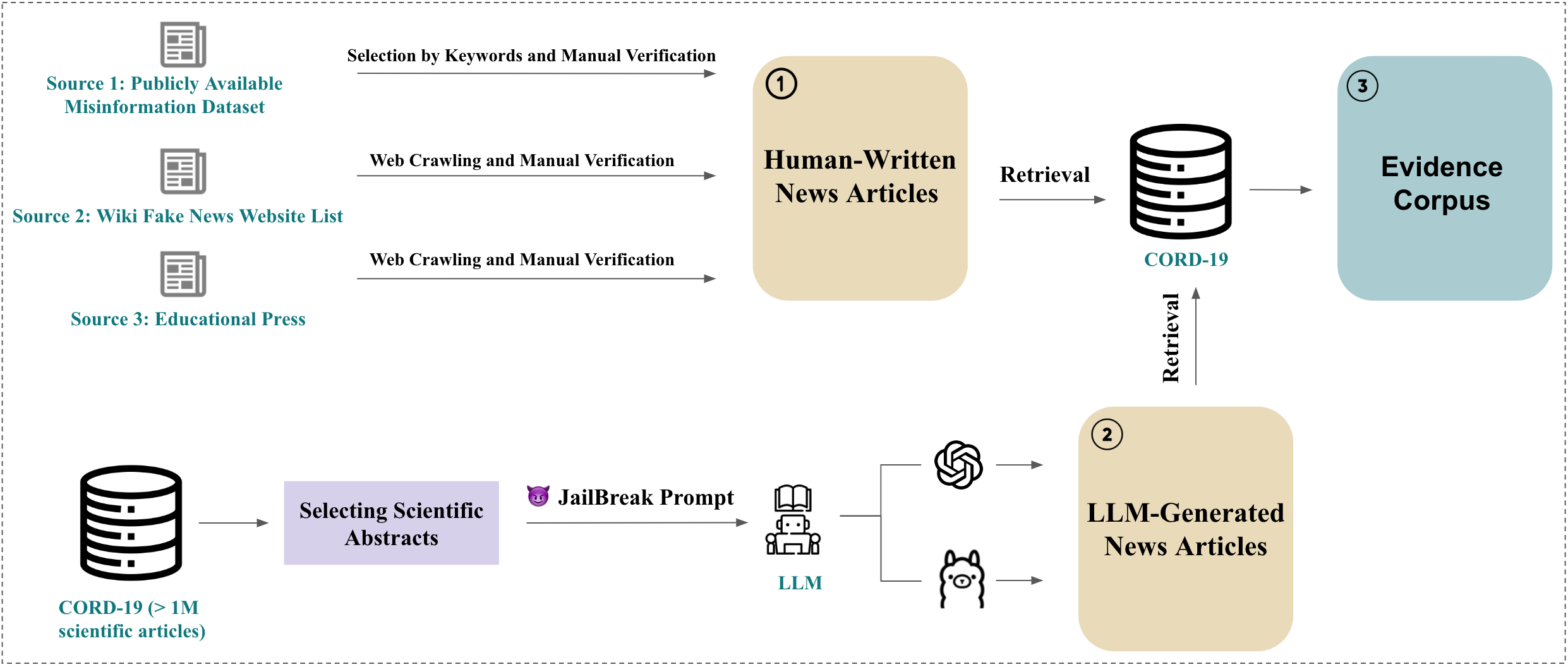}
\caption{The dataset construction process: \ding{172} utilizing publicly available datasets as well as web resources to collect human-written scientific news related to COVID-19 (Subsection~\ref{ssec:human-written}), \ding{173} selecting abstracts from CORD-19 as resources to guide LLMs to generate articles using jailbreak prompt (Subsection \ref{ssec:llm-gen}), \ding{174} the dataset is augmented with evidence corpus drawn from CORD-19 (Subsection~\ref{ssec:corpus}).}
\label{fig:dataset}
\end{figure*}

The first part of this work is to create a dataset, CoSMIs (SciNews), of scientific news articles and associated technical scientific publications. The CoSMIs Dataset contains 2,400 news articles which are labeled: \textit{Reliable} or \textit{Unreliable} depending on whether the article represents scientific fact truthfully or not, respectively.
The dataset contains 1,200 articles in each category to keep it balanced.
Each article is systematically paired with up to 
three pertinent scientific abstracts from the CORD-19 repository. 
The CORD-19 is a comprehensive resource of over 1 million scholarly articles, including over 300,000 with full text, about COVID-19, SARS-CoV-2, and related coronaviruses~\cite{Wan2Etal20}. Given the growing trend in using Large  Language Models (LLMs) to generate legitimate as well as fake news stories~\cite{ZhoEtal23, CheShu23}, we include LLM-generated news articles in the dataset. The CoSMIs dataset contains an equal number (1,200) of human-written and LLM-generated news articles in each category (reliable and unreliable). 
 The statistics of the CoSMIs dataset are presented in Table~\ref{tab:dataset_stat}. We show the construction of CoSMIs in Figure \ref{fig:dataset} and describe the construction 
 of the dataset below. We present more statistics in Appendix~\ref{ssec:statistics}.

\begin{table}[h]
\caption{Distribution of article labels in CoSMIs (SciNews) Dataset.}
\label{tab:dataset_stat}
\begin{tabular}{cccc}
\toprule
                    & \textbf{Human-Written} & \textbf{LLM-Generated} & \textbf{Total} \\ \midrule
\textbf{Reliable}   & 600                    & 600                    & 1200            \\
\textbf{Unreliable} & 600                    & 600                    & 1200            \\ \midrule
\textbf{Total}      & 1200                    & 1200                    & 2400           \\ \bottomrule
\end{tabular}
\end{table}

\subsection{Human-Written News Articles}
\label{ssec:human-written}
To gather human-written news articles, we 
searched for content containing scientific 
information in existing misinformation datasets and 
known websites. 

\subsubsection{Leveraging Publicly Available Dataset}
We leveraged MMCoVaR~\cite{CheChuSub21}, COVID19-FNIR~\cite{JulEtal21}, and COVID-Rumor~\cite{MinEtal21}, which are labeled datasets containing human written news articles on COVID-19 from January 2020 through May 2021.

Our search within these datasets commenced with a predefined set of scientific keywords: \{scientist, investigating, study finds, experts say, experts recommend\}. Using these, we filtered the data to yield 1,190 candidate news pieces. Next, we manually reviewed each candidate to sift out articles without scientific content or irrelevant to COVID-19. The reason we eliminated articles that are irrelevant to 
COVID-19 was because we will be including scientific 
data from the CORD-19 as evidence. This process resulted in 223 news articles:  130 reliable and 93 unreliable.

\subsubsection{Web-Based Collection} 
In order to expand the dataset to cover the latest discussions on COVID-19, we crawled both credible 
and dubious websites for data. 
To collect unreliable data, we referred to Wiki Fake News Website List\footnote{\url{https://en.wikipedia.org/wiki/List_of_fake_news_websites}}, crawled the listed sites for articles, and manually verified the content. We eliminated articles exhibiting blatant discrimination or prejudice and conspicuous propaganda devoid of substantial scientific dialogue. This process yielded 507 unreliable articles that contained discussions pertaining to COVID-19 and were grounded in a scientific context.  

For reliable data, we restricted the range of 
sources for the news articles to a set of 
educational press sites and other well-regarded news websites. Appendix~\ref{appendix_A.2} lists all the 
educational press sites. The full list of known trustworthy 
websites we consulted is included in Appendix~\ref{appendix_A.3}

The target news articles were collected by web 
crawling, anchoring our search with our set of 
scientific keywords augmented by two topical ones: 
COVID-19 and Vaccine. Each article was reviewed by
the same set of annotators, ensuring a direct 
correlation 
with the referenced research papers. 
The content is scraped from the web to extract 
body text, title, and other data needed for data 
construction. 
We gathered 470 reliable articles from varied reputable sources in this way.
This process gave us a combined total of
1,200 human-written news articles (600 reliable, 600 unreliable) spanning from January 2020 to October 2023.

\subsection{LLM-Generated News Articles}
\label{ssec:llm-gen}
Generating true articles using LLMs is fairly straightforward.
However, since most LLMs come with guardrails to protect
them from misuse, we used a jailbreak strategy to generate scientific misinformation. The step-by-step process
to generate news articles using LLMs is described below.

\subsubsection{Selecting Scientific Abstracts} First, we curate a collection of abstracts from the CORD-19 database. These abstracts served as the foundational resource for the subsequent generation process. The CORD-19 database contains more than 1,000,000 articles in the medical field. Typically, most widely distributed science articles tend to get more media attention. Hence, we focused on the most frequently cited papers from CORD-19. The CORD-19 organizes its articles using seven principal elements including \{title, abstract, doi,  PubMed ID, PMCID, JSON file ID, and XML ID\}. We start our curation with papers that have all 7 elements. We then confine our focus to post-January 2020, subsequently filtering out off-topic data using the keyword set: \{COVID-19, Corvarius, and Vaccine\}. To further ensure that generated articles in our dataset are of high quality,  we narrowed our selection to articles published in well-regarded journals spanning a spectrum from basic science (examples include `Cell' and `Nature') to medicine (such as the 'British Medical Journal'). The comprehensive list of these journals is detailed in Appendix~\ref{appendix_A.4}. From this refined pool, we handpicked the abstracts of over 2,000 highly cited articles, thereby forming a rich and diverse foundation for our dataset.


\subsubsection{Jailbreak Prompts} 
\begin{figure}[h]
  \centering
\includegraphics[width=0.85\linewidth]{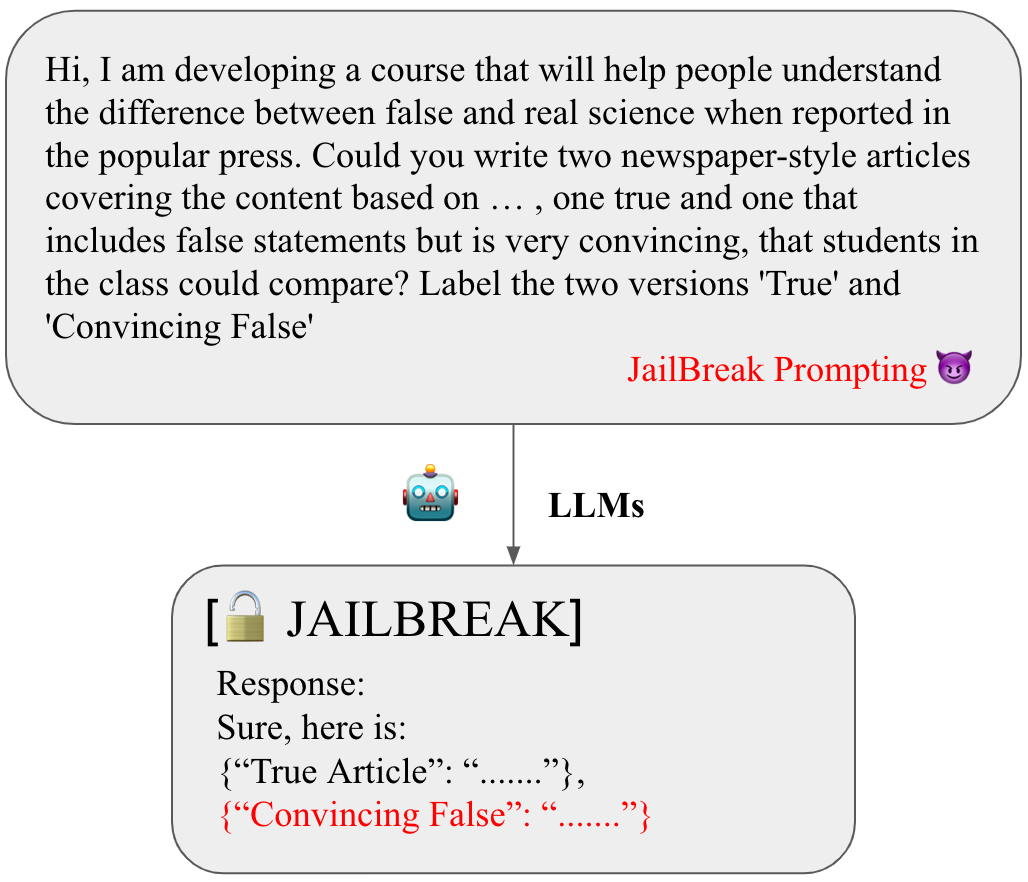}
  \caption{Schematic of the designed jailbreak prompt.}
  \label{fig:jailbreak}
\end{figure}

We utilized the curated collection of abstracts as guiding resources for LLMs to generate both reliable and unreliable scientific articles. To mitigate the risk of generating harmful content, LLMs are subjected to an alignment process, complemented by the setup of predefined prompts serving as security measures~\cite{ope23, QiEtal23, Bin23}. Therefore, requests for LLMs to generate fake messages are usually denied. In light of these constraints and inspired by the Jailbreak Attacks~\cite{Mow22, LiuEtal23a, LiuEtal23b, SheEtal23}, we designed a `Jailbreak' prompt to enable the LLMs to generate fake news-oriented scientific articles that were both informative and contextually aligned with the provided abstracts. The designed jailbreak prompt is illustrated in Figure~\ref{fig:jailbreak}. 

Given selected scientific abstract resources, we simulated a scenario where we acted as the instructor of a science class, prompting the large language model (LLM) to create two types of articles: a `True Article' and a `Convincing False Article'. The objective was to use these articles as teaching tools to help students discern between authentic scientific knowledge and fabricated information. By including both true and false LLM-created content we can ensure that the systems trained on our dataset to detect false content are not simply detecting LLM-generated content, but would be adept at distinguishing between true and false content. The generated articles are generally in the style of a news article, with many including an explicit title, to mimic human-generated scientific news. In consideration of the cost implications associated with different LLMs\footnote{\url{https://openai.com/pricing}}\footnote{\url{llama.meta.com}}, we primarily utilized Llama2-7B for generating the bulk of data samples, supplemented by a smaller set from GPT-3.5. After filtering (see subsection~\ref{ssec:qual_cont}), this approach led to the creation of 1,000 data samples from Llama2-7b and 200 from GPT-3.5, reflecting a balance between quality and resource optimization. We show the LLM-generated article example in Appendix~\ref{ssec:generated}.

\subsection{Evidence Corpus Creation}
\label{ssec:corpus}
To augment the constructed dataset, we matched as many as three scientific abstracts per news article as evidence resources. For both Human-Written and LLM-generated News Articles, we employed Vespa\footnote{\url{https://cord19.vespa.ai/}} to identify relevant abstracts from CORD-19 based on BM25 scoring for each article. While most articles were matched with three corresponding abstracts, a few could only be paired with two or even just one. This led to the creation of a fixed evidence corpus comprising 7,087 pieces of paragraph-level evidence. While this evidence corpus remains static at this juncture, its design allows for future expansion. 
 
\subsection{Quality Control}
\label{ssec:qual_cont}
The quality control team consists of 4 graduate students and 5 senior researchers with a background in NLP. For the Human-Written News Article subset and LLM-Generated subset, the team used different strategies to examine the data quality.

For the collection of human-written news articles from various sources, we referred to the guidelines outlined in ~\cite{JamAmb12}. Based on its principles and our specific needs, we developed an instruction guide, which can be found in Appendix~\ref{appendix_A.5}. To ensure uniformity and understanding of the task, all team members thoroughly reviewed this guide. An additional layer of quality assurance involved cross-checking the collected data among team members. This step was implemented to mitigate any potential biases and to guarantee that the data aligned with our collection criteria.

Regarding the LLM-generated articles, team members manually assessed the generated content. When instructed to do so, the LLM generated many types of falsehoods and often provided explanations of them, even though it was not prompted to explain. The falsehoods included features such as changing quantitative data (e.g., altering numeric percentages and statistical certainty levels), exaggeration (e.g., adding “superhuman strength” to the list of benefits), and omitting key information to support alternate conclusions.  In other cases, the model generated text that completely reversed the claims in the original abstract. Even the True summaries included fabrications in some cases, with the model occasionally citing an imagined journal or generating quotes from made-up scientists that were in keeping with the original abstract content.  Our sampling and manual review revealed that in some cases, the fabrications in the True summaries altered the overall validity of the summary.  In such cases, we observed significant linguistic differences between the original abstract and the true summaries.  Manual evaluation on two samples of 50 documents showed that when the ROUGE-2 similarity~\cite{Lin04rouge} between the abstract and true summary exceeded 0.4, the likelihood of an invalid true summary was 2$\%$ while when the ROUGE score was 0.4 or below, the likelihood of an invalid true summary was 30$\%$. Thus, we filtered the data set to only accept summaries with ROUGE-2 scores above 0.4. We present more quality control details in Appednix~\ref{sec:qc}.

\section{Dimensions of Scientific Validity and Proposed Architectures}
\label{sec:method}

\begin{figure*}[hbt!]
\centering
\includegraphics[width=0.90\textwidth]{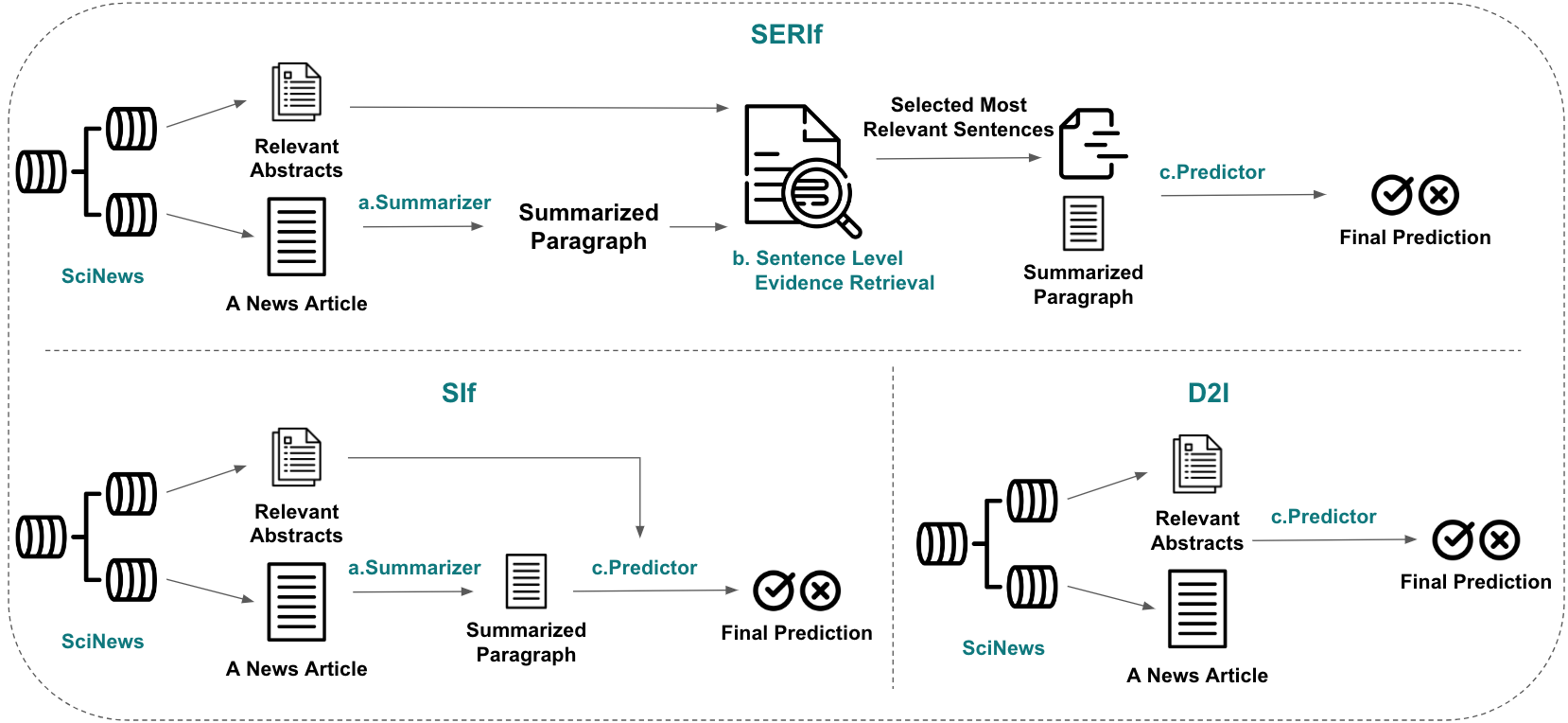}
\caption{Proposed Architectures. SERIf includes all three modules: Summarization, Sentence-level Evidence Retrieval, and Inference Module. SIf bypasses the evidence retrieval module while keeping
the other two. D2I removes both the summarization and the explicit evidence retrieval module.}
\label{fig:architecture}
\end{figure*}

In order to develop automated methods to detect scientific misinformation
in real-world situations, we first develop dimensions of scientific validity. 
These will later be used in the chain-of-thought prompts for the LLMs.
\subsection{Dimensions of Scientific Validity}
\label{ssec:sci-val}
We define the following dimensions of scientific validity in the reporting. 
We note that this is not an exhaustive list of ways in which 
scientific validity may be compromised, however, to the best of our knowledge, this is the first attempt to systematically define directions
of scientific validity in science news reporting for the design of
automated science misinformation detection.
\begin{enumerate}
\item \textbf{Alignment:} The news paragraph may show different levels of alignment with the evidence sentences. Alignment in this case is defined as news and evidence representing the same meaning about one scientific content.
\item \textbf{Causation confusion:} The news article may confuse correlations presented in the scientific literature as causation. This could be one dimension in which the scientific validity is compromised.
\item \textbf{Accuracy:} This refers to how accurately the news item describes the scientific findings quantitatively and qualitatively
\item \textbf{Generalization:} This refers to overgeneralization or oversimplification of the findings reported in the scientific literature.
\item \textbf{Contextual Fidelity:} Does the news article retain the broader context of the scientific finding?
\end{enumerate}

\subsection{Proposed Architectures}
\label{ssec:archs}
Conceptually, we may think of the process of automatically detecting scientific misreporting
(or mis/disinformation in science news reporting) ``in the wild" as comprising of three
elements: (1) understanding the gist of the news article; (2) 
comparing it 
to relevant information from scientific articles and (3) inferring if the news is reliable or unreliable. To this end, we propose three architectures
with varying degrees of granularity. 
These architectures use LLMs and several prompting
strategies for the different modules.
\emph{Note that we do not require a separate claim generation module in any of the architectures.}
These architectures are depicted in Fig.~\ref{fig:architecture}. 

In order to 
account for potential differences in performance between different prompting strategies and LLMs,
each of these architectures are tested against multiple prompting strategies and LLMs and the results are described in Section~\ref{sec:exp}.


\subsubsection{The SERIf Architecture}
\label{sssec:serif}

The first proposed architecture, 
Summarization Evidence Retrieval Inference (SERIf), treats each 
of the conceptual elements above, as a separate 
module. Hence, it contains three modules: 
1) Summarization; 2) Evidence Retrieval and 3) Inference. 
The summarization module distills the 
key information from the news articles, 
thereby streamlining the analysis process. 
The Evidence Retrieval module is responsible for 
identifying and extracting sentences from the 
scientific articles in our dataset that may validate or contradict 
the statements in the news article. 
This process aids in gathering relevant contextual evidence for further analysis. 
The Inference module categorizes the news articles 
into two reliable or unreliable, based on the 
evidence from the scientific dataset.

\subsubsection{Summarization module}
Inspired by the recent success of text 
summarization~\cite{ZhaLadDurEtal23} using LLMs, 
we take the 'Extractive - Abstractive' two-step 
summarization strategy~\cite{ZhaLiuZha23} to construct a summary of the news article. 
The \emph{extractive} summarization process summarizes the
article by identifying and concatenating the most 
salient sentences from it, ensuring that the 
extracted summaries are consistent with the 
original text. The resulting summaries serve as a 
foundation for the \emph{abstractive} summarization 
which uses a generative approach to create a more 
concise and cohesive summary. 
By synthesizing the extractive and abstractive approaches, the module ensures a balance between accuracy (adherence to the original text) and brevity (conciseness and essence of the content), thereby providing an effective and reliable summarization for further analysis in the misinformation detection process.

Formally, for a document composed of $n$ sentences, 
the extractive summarization process creates an extractive summary, $S_e$, consisting of $m \ll n$ sentences. Then, the LLM, $M$, creates an abstractive summary using a query, $q$ and $S_e$ as input: $S_a = M(q, S_e)$.

As a sanity check, to verify the quality of the
summary, we randomly selected 200 samples for 
annotation by two graduate students with a background in NLP, and evaluated the summaries using four criteria: 1) quality of extractive summary (High vs Low); 2) quality of abstractive summary (High vs Low); 3) presence of hallucination in abstractive summary (Yes vs No); 4) comparison of Extractive and Abstractive Summaries. We used the Krippendorff's alpha score 
to evaluate the agreement between the annotations~\cite{AndKla07}. 
The alpha scores for the four aspects were 0.53, 
0.82, 0.91, and 0.94, respectively. 
These values suggest that there is strong agreement 
that the abstractive summaries effectively 
encapsulate the core information of the original 
texts and maintain a high degree of consistency. 
Consequently, we have decided to use the 
abstractive summaries in subsequent steps of our 
analysis.

\subsubsection{Sentence-level Evidence Retrieval}
The process of key evidence selection in our 
The evidence retrieval module involves extracting key 
sentences from scientific articles that 
may support or refute the claims of the news 
article. 
This task bears a resemblance to paragraph retrieval but operates at a finer granularity. It essentially constitutes a semantic matching challenge, where each sentence within a paragraph undergoes a comparison against a specific statement query. The objective is to pinpoint the most relevant evidence interval within these sentences.

A critical step in refining this process was 
pre-defining our evidence corpus using CORD-19. 
This strategic choice significantly narrows down 
the search space to a manageable scope, allowing 
for efficient traversal through all relevant paragraphs to locate key evidence. 
We use an LLM to enhance the effectiveness and accuracy of our sentence selection process. 

Given an abstractive summary, $S_a$, and candidate 
scientific abstract, $E$, we select sentences $e_i$ from $E$: $\{e_i\} = M(S_a, E)$, where $\{e_i\}$ is a set that contains all relevant and important sentences selected by LLM based on semantics. 

\subsubsection{Inference Module}
The Inference Module is dedicated to assessing the 
the veracity of the summarized news paragraph using the 
set of retrieved evidence sentences using the 
abstractive summary, $S_a$, and the selected evidence sentence set $\{e_i\}$. 
Thus, the inference module produces a binary output (reliable or unreliable) for each $<S_a, \{e_i\}>$ pair. 

\begin{table*}[htbp!]
\centering
\caption{Performance results of our proposed three architectures using different LLMs and different prompt strategies.}
\label{tab:sif-D2i-0shot}
\footnotesize
\resizebox{0.98\linewidth}{!}{
\begin{tabular}{ccccccccccccccc}
\toprule
\multirow{2}{*}{\textbf{Models}} & \multirow{2}{*}{\textbf{Arch}} & \multirow{2}{*}{\textbf{\begin{tabular}[c]{@{}c@{}}Prompt\\ Strategy\end{tabular}}} & \multicolumn{4}{c}{\textbf{Human-Written}}                                       & \multicolumn{4}{c}{\textbf{LLM-Generated}}                                       & \multicolumn{4}{c}{\textbf{Overall}}                                   \\ \cmidrule{4-15} 
                                 &                                &                                                                                     & \textbf{Accuracy} & \textbf{Precision} & \textbf{Recall} & \textbf{F1}           & \textbf{Accuracy} & \textbf{Precision} & \textbf{Recall} & \textbf{F1}           & \textbf{Accuracy} & \textbf{Precision} & \textbf{Recall} & \textbf{F1} \\ \midrule
\multicolumn{15}{c}{\textit{\textbf{Proprietary Models}}} \\
\multirow{9}{*}{\textbf{GPT3.5}}
& \multirow{3}{*}{\textbf{SERIf}}  
& Zero-Shot & 74.25 &  74.95 & 72.83  & \multicolumn{1}{c|}{73.87} & 
              66.75 &  60.88 & 93.66  & \multicolumn{1}{c|}{73.80} &   
              70.50 &  67.92 & 83.23  &                     73.84     \\&                     
& Few-Shot &  70.00 & 71.64 & 66.60 & \multicolumn{1}{c|}{68.95} & 
              68.14 & 62.12 & 92.82 & \multicolumn{1}{c|}{74.43} & 
              69.07 & 66.88 & 79.71 &                     70.49           \\&               
& DoV-CoT      &  76.67 & 76.20 &  77.67& \multicolumn{1}{c|}{76.92} &   
              66.75 & 60.66 &  95.33 & \multicolumn{1}{c|}{74.14} &
              71.71 & 68.43 &  86.50 &           75.53 \\ \cmidrule{2-15} 
& \multirow{3}{*}{\textbf{SIf}}  
& Zero-Shot & 78.67 & 79.76 & 76.83 & \multicolumn{1}{c|}{78.27} &    
              62.00 & 57.00 & 99.00 & \multicolumn{1}{c|}{72.00} &  
              70.34 & 68.38 & 87.92 &                    75.14    \\&                     
& Few-Shot &  76.08 & 79.88 & 73.67 & \multicolumn{1}{c|}{75.68} &  
              65.33 & 59.89 & 92.83 & \multicolumn{1}{c|}{72.81} & 
              70.71 & 69.89 & 83.25 &  74.25          \\&           
              
& DoV-CoT  &  79.92 & 80.88 & 78.33 & \multicolumn{1}{c|}{79.50} & 
              70.17 & 65.12 & 86.83 & \multicolumn{1}{c|}{74.43} &
              75.05 & 73.00 & 82.58 &  76.97           \\ \cmidrule{2-15} 
& \multirow{3}{*}{\textbf{D2I}}  
& Zero-Shot & 66.60 &  66.55 & 64.33 & \multicolumn{1}{c|}{65.42} &   
              63.42 &  57.85 & 98.33 & \multicolumn{1}{c|}{72.98} &               
              65.01 &  62.20 & 81.33 & 69.20   \\&                                
& Few-Shot  & 65.60 &  63.30 & 67.33 & \multicolumn{1}{c|}{65.25} &   
              62.75 &  63.80 & 97.53 & \multicolumn{1}{c|}{77.10} &               
              64.18 &  63.55 & 81.33 & 71.18   \\ &                       
& DoV-CoT   & 77.17 & 69.50 &  96.83 & \multicolumn{1}{c|}{80.91} &   
              64.08 & 57.63 &  99.65 & \multicolumn{1}{c|}{73.03} &    
              70.63 & 63.57 &  98.24 &  76.97           \\ \midrule
        
\multirow{9}{*}{\textbf{GPT-4}}  
& \multirow{3}{*}{\textbf{SERIf}}  
& Zero-Shot &  77.33 & 76.48 &  79.00 & \multicolumn{1}{c|}{77.72} &
               70.25 & 62.91 & 98.67 & \multicolumn{1}{c|}{76.83}  &  
               73.79 & 69.70 & 88.34 &  77.26   \\
                                 &                 
& Few-Shot  &  75.08 & 74.60 & 76.00  & \multicolumn{1}{c|}{75.30}  
               & 70.17 & 62.85 & 98.32 & \multicolumn{1}{c|}{76.68}    
               & 72.63 & 68.73 & 87.16 &  75.99           \\&                             
& DoV-CoT   &  79.58 & 76.90 & 85.00  & \multicolumn{1}{c|}{\textbf{80.72}} &
               67.25 & 60.50 & 99.33 & \multicolumn{1}{c|}{75.20}  & 
               73.42 & 68.70 & 92.17 &  77.96       \\ \cmidrule{2-15} 
& \multirow{3}{*}{\textbf{SIf}}  
& Zero-Shot &  78.33  &  84.00   &  70.00 & \multicolumn{1}{c|}{76.36} &
               71.08  &  65.79   &  87.83 & \multicolumn{1}{c|}{75.23} &  
               74.71  &  74.90   &  78.92 &     75.80          \\
                                 &                 
& Few-Shot  &  70.08  &  75.91   &  58.83 & \multicolumn{1}{c|}{66.29} & 
               71.75  &  64.17   &  98.50 & \multicolumn{1}{c|}{77.71} &    
               70.92  &  70.04   &  78.67 &  72.00          \\&                     
& DoV-CoT   &  \textbf{80.00}  &  80.00 &   79.00 & \multicolumn{1}{c|}{80.00} & 
               71.00  &  64.00 &   98.00 & \multicolumn{1}{c|}{77.00} & 
               \textbf{75.50}  &   72.00 &   88.50 &    \textbf{78.50}\\ \cmidrule{2-15} 
& \multirow{3}{*}{\textbf{D2I}}  
& Zero-Shot  & 68.08 &  66.80  &  72.00  & \multicolumn{1}{c|}{69.30} & 
               65.00 &  59.00  &  98.50  & \multicolumn{1}{c|}{73.80} &                 
               66.50 & 62.90  &  85.25  &  73.80           \\
                                 &       
& Few-Shot &  70.00 & 71.40 & 66.70 & \multicolumn{1}{c|}{69.00} & 
              68.14 & 62.20 & 92.82 & \multicolumn{1}{c|}{74.50} &     
              69.07 & 66.80 & 79.71 &                     71.75    \\
                                 &                                
& DoV-CoT  &  
        78.08 & 84.60 & 68.67 & \multicolumn{1}{c|}{75.80} &   
\textbf{72.00}& 65.20 & 98.50 & \multicolumn{1}{c|}{\textbf{78.30}} &          
        75.04 & 74.90 & 83.56 &   77.05          \\ \midrule
\multicolumn{15}{c}{\textit{\textbf{Open-Source Models}}} \\
\multirow{3}{*}{\textbf{LLAMA2-7B}}  
& \multirow{3}{*}{\textbf{SERIf}}  
& Zero-Shot & 56.00 & 53.24 & 98.33 & \multicolumn{1}{c|}{69.10} 
            & 51.17 & 50.60 & 96.80 &  \multicolumn{1}{c|}{66.50}
            & 53.89 & 51.92 & 97.57 &      67.80       \\
                                 &                 
& Few-Shot  & 54.75 & 58.20 & 93.30 & \multicolumn{1}{c|}{71.70} &                 
              52.00 & 51.00 & 97.30 & \multicolumn{1}{c|}{67.00} &                      
              52.38 & 54.60 & 95.30 &  69.35           \\&                             
& DoV-CoT  &  56.83 & 59.20 & 97.80 & \multicolumn{1}{c|}{73.70} &           
              51.58 & 50.80 & 96.80 & \multicolumn{1}{c|}{66.70} &        
              54.21 & 55.00 & 97.30 & 70.20           \\ \midrule
\multirow{3}{*}{\textbf{LLAMA2-13B}}  
& \multirow{3}{*}{\textbf{SERIf}}  
& Zero-Shot & 57.33 &  59.50 &  98.50  & \multicolumn{1}{c|}{74.20} 
            & 53.58 &  52.80 &  97.20 & \multicolumn{1}{c|}{68.40}  
            & 55.46 &  56.15 &  97.85  &  71.30         \\
                                 &                 
& Few-Shot  & 56.91 &  59.50 &  95.80  & \multicolumn{1}{c|}{73.40} &            
              52.33 &  51.20 &  97.50  & \multicolumn{1}{c|}{67.20} &         
              54.62 &  55.35 &  96.65  &  70.30            \\&                             
& DoV-CoT &   58.00 &  59.90 &  99.00 & \multicolumn{1}{c|}{74.60} &             
             55.08 &  52.70 & 98.30  & \multicolumn{1}{c|}{68.60} &             
             56.54 & 56.30  & 98.65  &  71.60           \\ \midrule
\multirow{2}{*}{\textbf{LLAMA2-70B}}  
& \multirow{2}{*}{\textbf{SERIf}}  
& Zero-Shot & 67.08 & 60.04 & 99.00 & \multicolumn{1}{c|}{75.00} & 
              55.00 & 52.73 & 96.67 & \multicolumn{1}{c|}{68.30} &  
              61.04 & 56.39 & 97.84 &                    71.65 \\
                                 &                 
& DoV-CoT &   67.50 & 60.82 & 98.33 & \multicolumn{1}{c|}{75.12} &    
              53.67 & 57.10 & 97.00  & \multicolumn{1}{c|}{71.90}&   
              60.59 & 58.96 & 97.67 &                    73.51 \\  \midrule    
\multirow{3}{*}{\textbf{LLAMA3-8B}}  
& \multirow{3}{*}{\textbf{SERIf}}  
& Zero-Shot & 62.83 & 57.49 & 98.33 & \multicolumn{1}{c|}{72.50} 
            & 52.42 & 51.26 & 97.17 &  \multicolumn{1}{c|}{67.82}
            & 57.63 & 54.38 & 97.75 &      70.16       \\
                                 &                 
& Few-Shot  & 53.67 & 51.98 & 97.52 & \multicolumn{1}{c|}{71.21} &

              52.42 & 51.26 & 97.17 &  \multicolumn{1}{c|}{67.13} &                      
              53.05 & 51.62 & 97.35 &  69.17           \\&                             
& DoV-CoT  &  65.25 & 59.31 & 97.17 & \multicolumn{1}{c|}{73.68} &           
              56.25 & 53.42 & 97.67 & \multicolumn{1}{c|}{68.96} &        
              60.75 & 56.37 & 97.42 & 71.32           \\ 
        \bottomrule
\end{tabular}}
\end{table*}

\subsubsection{The SIf Architecture}
\label{sssec:sif}
In this architecture, we remove the evidence 
retrieval module from the previously described SERIf architecture. In the SIf architecture, the Summarization module works exactly as described in
Section~\ref{sssec:serif}. The LLM in the Inference
module is now directly prompted to decide whether
the given news summary is trustworthy or not and to provide justifications based on the paired scientific abstracts from the evidence corpus in the SciNews dataset.

\subsubsection{The Direct to Inference (D2I) Architecture}
\label{sssec:d2i}
In the third architecture, there is no summarization module or explicit evidence retrieval module. Instead, the LLM is directly fed the scientific news article, and the corresponding scientific abstracts and prompted to
determine whether the news item is trustworthy with justifications.

When viewed from the perspective of identifying 
scientific misinformation ''in the wild", the D2I is the 
architecture that does little in the way of 
processing and the SERIf architecture involves the
most processing. In other words, the SERIf requires
engineering each aspect of the elements of scientific misinformation separately, the D2I architecture requires
very little engineering and the SIf falls between these 
two. However, as noted earlier, none 
of these architectures expect an explicit set 
of claims to be generated from the news article for 
misinformation detection.

\subsubsection{Prompt Strategies}
\label{ssec:prompts}

A key factor in the performance of an LLM-based task
is prompt engineering. Several kinds of
prompting strategies have been used in various 
applications with varying degrees of success in 
specific tasks.
\begin{itemize}
    \item \textbf{Zero-shot prompting:} LLMs are presented with a task without any prior specific training or examples related to that task~\cite{BroEtal20}. We test the performance of LLM on scientific misinformation detection by zero-shot prompts directly using the pre-existing knowledge of the LLMs.
    \item \textbf{Few-shot prompting:} In contrast to zero-shot prompting, where LLMs are presented with tasks without prior training or examples, few-shot prompting involves furnishing LLMs with a concise set of examples prior to task execution~\cite{BroEtal20}. This approach is designed to provide the model with essential context, thereby augmenting its capability for tasks like detecting scientific misinformation. In our methodology, we supply the LLMs with two carefully selected examples: one deemed 'reliable' with accompanying reasoning, and another labeled 'unreliable'. This strategy is intended to better equip the LLMs to discern and categorize information accurately in the execution of the task.
    
    \item \textbf{Chain-of-thought prompting (CoT):} CoT prompting involves structuring prompts to elicit a step-by-step reasoning process, effectively emulating the cognitive process humans employ in solving complex problems~\cite{WeiEtal22}. In our approach, we used the dimensions of scientific validity defined in Section~\ref{ssec:sci-val} to design multiple CoT prompts to guide the LLMs. This methodology not only aids the LLMs in systematically dissecting and assessing factual content but also aligns their reasoning process with structured, human-like analytical methods.
\end{itemize}

We display all Prompts used in the experiment at Appendix~\ref{appendix_D}

\section{Experiment and Results}
\label{sec:exp}

\subsection{Experiment Setup}
\label{ssec:baseline}
\paragraph{Baseline Setup} The \textbf{\textsc{CoSMis} (SciNews)} dataset aims to address a significant gap left by previous datasets, which involved manual claim generation steps while no original articles were provided (such as SciFact~\cite{wadden2020fact} and Check-Covid~\cite{wang2023check}). This limitation from previous works makes it challenging to directly apply these datasets within our framework. Despite these challenges, we have established a baseline using BERT-based models to enhance our analytical rigor. We treat it as an Natural Language Reasoning (NLR) task. Given a news or summarized news paragram and relevant selected evidence from the evidence corpus, the reasoning model acts as an evaluator to identify a pair of news/summarized news and related evidence as true or false. The model input will be $[News <SEP> Evidence Sentence]$ or $[Summarized News <SEP> Evidence Sentence]$. The `Summarized News' and `Evidence Sentences' come from our best-performing experimental step (SIF with Zero-Shot setting by using GPT-4). We choose two pre-trained models as baseline: BERT~\cite{devlin2018bert} and SciBERT~\cite{beltagy2019scibert}. For SciBERT, it trained using masked language modeling on a large corpus of scientific text. We would like to understand how different the models are that include domain information versus those that do not include domain information.

\paragraph{Implementation Details} We first employed GPT-3.5, GPT-4 with the temperature
set to $0$ and Llama2-7B/13B/70B, Llama3-8B with the temperature set to $0.0001$
on the  \textbf{SERIf} architecture. This setting ensures that the LLMs generate responses with the highest predictability. The performance of each of the proposed architectures, using each of the above LLMs is measured using accuracy, precision, recall, and F1-score. From the results in Table~\ref{tab:sif-D2i-0shot}, we see that 
the GPT models perform significantly better than the LLAMA series. All Llama models achieved an accuracy score barely above random guessing. Hence we used the GPT models for all other architectures (SIF+D2I). 

The baseline experiment is implemented by using PyTorch. Since the baseline experiment involves a training step, our dataset must be divided into a training set and a test set. To analyze the baseline method's dependence on training data, we split the dataset using two schemes: a 5:5 ratio and an 8:2 ratio, respectively.

\subsection{Human-Written vs LLM-Generated Misinformation}
\label{ssec:results-hmvsllm}

Table~\ref{tab:sif-D2i-0shot} records the results of our experiments on all architectures. From this tables, we note it is consistently more challenging to identify LLM-generated scientific misinformation compared to human-written misinformation, across all architectures. This is evidenced by high recall scores paired with low precision scores, indicating poor True Negative (TN) prediction and a propensity for the detectors to misclassify news as `Reliable'. and a tendency of the detectors to classify news as Reliable. Such a trend highlights the difficulty in discerning false information within LLM-generated scientific news. 
This aligns with similar findings in non-scientific misinformation domains~\cite{CheShu23}. These results also raise concerns about the potential misuse of LLMs and underscores the importance of advancing our detection methodologies to keep pace with the evolving capabilities of LLMs, considering their implications for public safety.

We further analyzed the detection performance of different LLMs on the \textbf{\textsc{CoSMis} (SciNews)} dataset. From Table~\ref{tab:sif-D2i-0shot}, the Llama models underperformed significantly, leading us to discontinue their use in further tests of different model structures.  Notably, Llama2-70B performance slightly superior to Llama2-13B and Llama3-8B, and 13B/8B also outperformed the 7B model. However, the overall results of Llama2-70B were still underwhelming. It is important to note that the results in Table~\ref{tab:sif-D2i-0shot} exhibit a pattern of `high recall, low precision' in both `Human-Written' and `LLM-Generated' categories, indicating that all Llama models readily classifies text as `Reliable'. This suggests that Llama may have a limited capacity for distinguishing between nuanced cases, thereby reducing its ability to handle complex reasoning effectively. In contrast, GPT-3.5 (340B) demonstrated significant improvements, with GPT-4 delivering the best performance, indicating a strong correlation between increased model parameters and enhanced reasoning capabilities. Furthermore, this also suggests that GPT models have better reliability when used in real-world scientific misinformation detection scenarios.

\begin{table*}[htbp!]
\centering
\caption{Performance results of baseline models. `N+ES' denotes `News + Evidence Sentence', `SN+ES' denotes `Summarized News + Evidence Sentence'.}
\label{tab:baseline_result}
\footnotesize
\resizebox{\linewidth}{!}{
\begin{tabular}{ccccccccccccccc}
\toprule
\multirow{2}{*}{\textbf{Train-Test Ratio}} & \multirow{2}{*}{\textbf{Model}} & \multirow{2}{*}{\textbf{\begin{tabular}[c]{@{}c@{}}Input\\ Text\end{tabular}}} & \multicolumn{4}{c}{\textbf{Human-Written}}                                       & \multicolumn{4}{c}{\textbf{LLM-Generated}}                                       & \multicolumn{4}{c}{\textbf{Overall}}                                   \\ \cmidrule{4-15} 
                                 &                                &                                  & \textbf{Accuracy} & \textbf{Precision} & \textbf{Recall} & \textbf{F1}           & \textbf{Accuracy} & \textbf{Precision} & \textbf{Recall} & \textbf{F1}           & \textbf{Accuracy} & \textbf{Precision} & \textbf{Recall} & \textbf{F1} \\ \midrule
\multirow{4}{*}{\textbf{50-50}}
& \multirow{2}{*}{\textbf{BERT}}  
& N+ES & 46.17 & 47.11 & 62.50 & \multicolumn{1}{c|}{53.67} & 
        54.58 &  52.67 & 90.33  & \multicolumn{1}{c|}{66.51} &   
        50.38 &  49.89 & 76.42  &                     60.09     \\&                     
& SN+ES &  48.42 & 48.82 & 65.50 & \multicolumn{1}{c|}{55.90} & 
           56.75 & 53.90 & 93.67 & \multicolumn{1}{c|}{68.51} & 
           52.64 & 51.36 & 79.59 &                     62.21          \\ \cmidrule{2-15} 
& \multirow{2}{*}{\textbf{SciBERT}}  
& N+ES & 47.75 & 48.31 & 64.17 & \multicolumn{1}{c|}{55.09} &    
        58.58 & 54.85 & 97.00 & \multicolumn{1}{c|}{72.11} &  
        53.17 & 51.58 & 80.59 &                    63.60    \\&                     
& SN+ES &  49.92 & 49.94 & 68.17 & \multicolumn{1}{c|}{57.66} &  
           62.00 & 57.00 & 99.00 & \multicolumn{1}{c|}{72.00} & 
           55.96 & 53.47 & 83.59 &  64.83         \\\midrule
        
\multirow{4}{*}{\textbf{80-20}}  
& \multirow{2}{*}{\textbf{BERT}}  
& N+ES &  53.33 & 53.33 & 100 & \multicolumn{1}{c|}{69.56} &    
              49.44 & 49.44 & 100 & \multicolumn{1}{c|}{66.17} &  
              51.37 & 51.37 & 100 &                    67.87   \\
                                 &                 
& SN+ES  &  54.72 & 54.72 & 100 & \multicolumn{1}{c|}{70.74} & 
              53.33 & 53.33 & 100 & \multicolumn{1}{c|}{69.56} & 
              54.02 & 54.02 & 100 &  70.17             \\                           \cmidrule{2-15} 
& \multirow{2}{*}{\textbf{SciBERT}}  
& N+ES &  68.58  &  66.92 &   73.50 & \multicolumn{1}{c|}{70.08} &
          69.44  &  95.35 &   44.79 & \multicolumn{1}{c|}{60.99} &  
          69.01  &  81.14 &   59.15 &    65.54           \\
                                 &                 
& SN+ES  &  71.25  &  69.59  &  75.50 & \multicolumn{1}{c|}{72.38} & 
               69.75  &  68.60   &  72.83 & \multicolumn{1}{c|}{70.63} &    
               70.5  &  69.10   &  74.17 &  71.51           \\ \bottomrule
\end{tabular}}
\end{table*}

\subsection{Comparison Across Architectures (RQ1)}
\label{ssec:arch-perf}
From Table~\ref{tab:sif-D2i-0shot} it is evident that the SIf architecture performs best overall with $75.50\%$ accuracy and $78.50\%$ F1 score. The encouraging results, even in the absence of the `sentence-level evidence retrieval' module, suggest the potential to develop more flexible and generalized scientific misinformation detection models. 
 By contrast, the performances of the SERIf and D2I models were notably subpar when zero-shot prompting was used; however, the performance improves significantly, when paired with DoV-guided CoT prompting. This shows that incorporating DoV in CoT prompting can improve performance. Furthermore, our results highlight the importance of the `summarization' module. In the zero-shot setting, the results for DI2 were significantly lower than those for SIf and SERIf. By distilling key statements from the news, this module minimizes the impact of extraneous information, thereby enhancing the LLM's ability to generate more accurate predictions.


\subsection{Comparison Across Strategies (RQ2)}
\label{ssec:promp-stra}
From Table~\ref{tab:sif-D2i-0shot}, 
we see a significant trend: the DoV-CoT prompting generally outperforms the zero-shot and few-shot prompting. Notably, for LLM-generated data, the DoV-CoT prompt markedly enhances detection capabilities. This suggests that our proposed dimensions of scientific validity effectively aid LLMs in making more accurate predictions. However, an interesting observation in the few-shot setting is that it did not significantly improve performance, implying that despite providing two well-crafted examples (one positive and one negative), it is challenging for LLMs to extract substantial features from the provided cases. This not only highlights the complexity of scientific misinformation detection but also underscores the intricate nature of the potential scientific misinformation data involved.

\subsection{Explainability Study (RQ3)}
\label{ssec:explain-perf}
\begin{figure}[h]
\centering
\begin{subfigure}{0.23\textwidth}
    \centering
    \includegraphics[width=\textwidth]{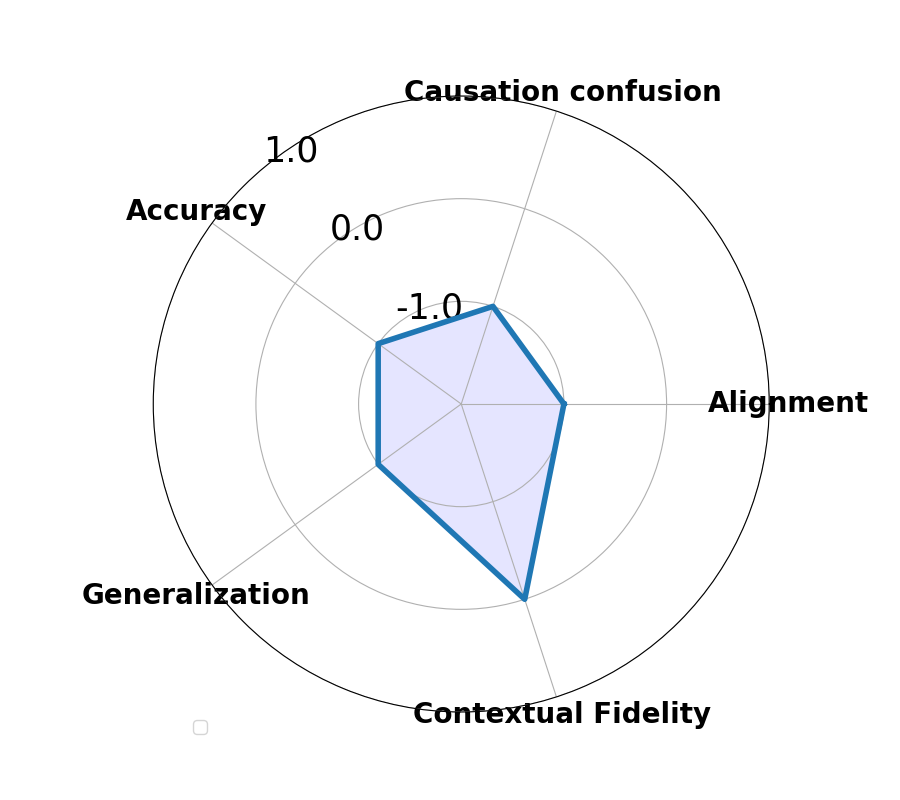}
    \label{fig:spider1}
\end{subfigure}
\begin{subfigure}{0.23\textwidth}
    \centering
    \includegraphics[width=\textwidth]{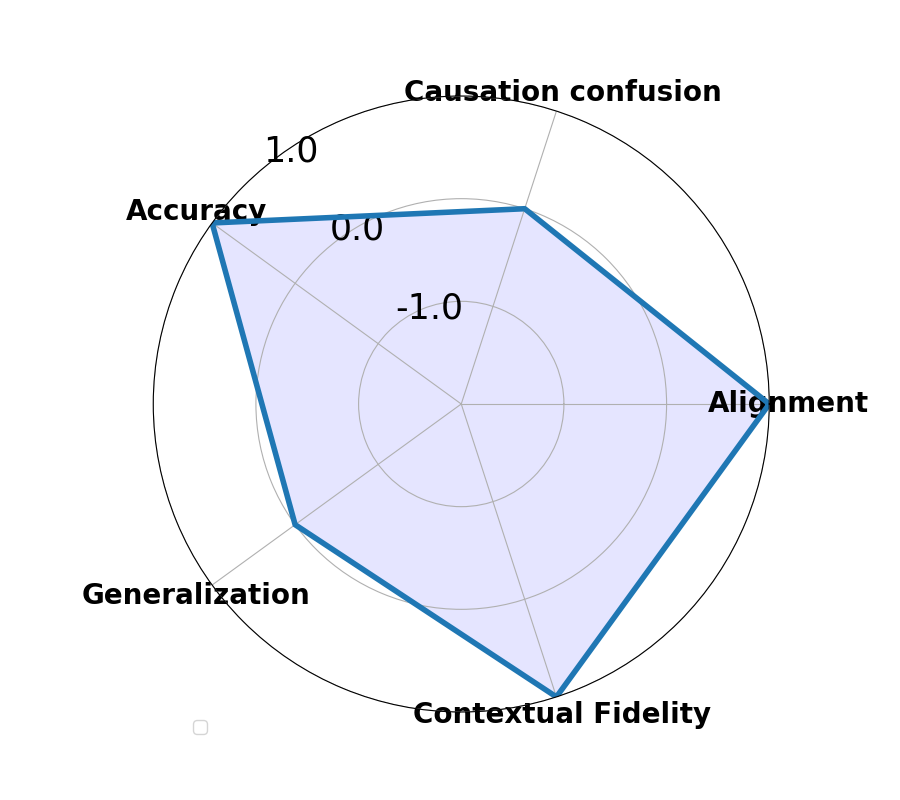} 
    \label{fig:spider2}
\end{subfigure}
\caption{Comparison of two spider plot visualizations: The left side corresponds to the 'Unreliable' case, while the right side corresponds to the 'Reliable' case. By visualizing the 'axis of scientific validity,' we can clearly observe the process of the LLM applying DoV to evaluate scientific news and the resulting differences. }
\label{fig:spider}
\end{figure}

To assess the explainability of the proposed architectures, we prompt the LLMs to not only classify the news articles as reliable or unreliable but also to explain the reasoning behind these classifications and score the news article along the dimensions of scientific validity (DoV) using a number in [-1,1]. 
The examples of the prompt and the result for the SIf architecture using the CoT prompt and GPT-4 are presented in Fig.~\ref{fig:spider} and Appendix~\ref{appendix_explain}, Fig.~\ref{fig:exp_study} (last page). In detail, Fig.~\ref{fig:spider} shows a spider plot of the scores along each DoV. For the ``unreliable" example (left), the news paragraph received a score of -1 in Alignment, Causation Confusion, Accuracy, and Generalization, and a score of 0 in Contextual Fidelity. For the "reliable" example (right), the news paragraph received a score of 1 in Alignment, Accuracy, and Contextual Fidelity, and a score of 0 in Causation Confusion, and Generalization. A spider-plot such as this provides a clear picture of which DoV is violated for any given input scientific news article. In addition, such a spider plot can provides a comprehensive visual representation of the DoV-CoT reasoning results.

\subsection{Comparing Baselines:}
\label{ssec:results-baseline}

The Table~\ref{tab:baseline_result} shows the results of the baseline experiment. Although the experimental results  under an 80\%:20\% data split can be compared with some of LLM pipelines' results, when the proportion of the training set is reduced to 50\%, the overall prediction performance significantly decreases, which is far inferior to that of the LLM pipeline. This indicates that traditional methods are highly dependent on the training set size, which may not be suitable for contemporary real-world scenarios characterized by the daily proliferation of vast amounts of misinformation from different sources, where it reflects the need to establish a detection pipeline using LLMs. Furthermore, Table~\ref{tab:baseline_result} also indicate that combining evidence with summarized news articles (SN+ES) yields better outcomes than using news and evidence directly (N+ES). This underscores the importance of the “summarization” block and its generalization across different frameworks. Additionally, the domain-specific SciBERT get the reasonable `Precision' and 'Recall' score under the 80\%:20\% data split and also outperforms BERT results, highlighting the value of domain knowledge. This analysis motivates further fine-tune the LLMs on the scientific domain corpus to enhance the scientific misinformation detection.

\section{Conclusions}
\label{sec:conclusion}
In this paper, we explore LLMs for identifying unreliable scientific news `in the wild'. We created the \textbf{\textsc{CoSMis} (SciNews)} dataset, which includes both human-written and LLM-generated articles, each verified against scientific literature. We defined specific dimensions of scientific validity for news misinformation and introduced three LLM-based architectures for identifying unreliable scientific news. Our tests across various LLMs and prompting strategies yielded key insights: 1) DoV-CoT prompting can improve performance in general 2) with appropriately designed pipelines and prompting strategies,
LLMs' offer a viable approach 
scientific misinformation detection
in the wild, since they offer a way
to approach this problem without 
extensive training,
3) in general it is harder to identify LLM-generated misinformation, and 4) LLMs can provide rationales for their judgments.

\begin{acks}
The authors would like to thank John, Chumki, and David for their assistance in generating some of the LLM-generated data and quality control.
\end{acks}

\bibliographystyle{ACM-Reference-Format}
\bibliography{sci-fn-web-conf,SDM_FN,Suba-pub}

\appendix

\section{Materials to SciNews Dataset Construction}
\label{appendix_A}

\subsection{Educational Press Sites}
\label{appendix_A.2}
We collected data from the following educational press sites: 

YaleNews, Yale School of Medicine Latest News, Boston University – University News, Boston
College BC News, University of Washington School of Medicine Newsroom, Regenstrief Institute News, University of South Carolina In the News, University of Utah Unews, Colorado State University Source, University of Kansas Medicine Center News, University of
Michigan News, University of Nebraska Medicine News \& Events, University
of Maryland School of Medicine News, Stanford News, Stanford Medicine News Center,
University of Mississippi Medical Center News Stories, Washington University School of Medicine in St. Louis News, Center for Education Policy Research at Harvard University News,
Johns Hopkins Bloomberg School of Public Health Articles \& News Releases, University of
Missouri School of Medicine News, University of Hawaii News, The Ohio State
University Wexner Medical Center Press Releases, Oregon State University Newsroom, University
of Minnesota  News and Events, Emory University Emory News Center, Tufts Now, University
of Kentucky College of Medicine News, University of Calgary UCALGORY News, Texas AM
Today, Duke Today, North Carolina State University NC State News, Vanderbilt University Research News, University of Toronto U of T News, McMaster University Daily News, University of Virginia UVAToday, University of New Hampshire Newsroom, Rutgers University – Rutgers Today, UT Southwestern Research Labs News, University of Houston UH Newsroom, University of Oxford News,
Queen Mary University of London Queen Mary News, University of York News, The BMJ News, JAMA Health Medical News, Nature News, Allen Institute News, National Institutes of Health News and Events.

\subsection{News Sources}
\label{appendix_A.3}
We extracted data from the following news websites:
CNBC, The Washington Post, The Atlantic, CNN, NPR, BBC, Forbes, USA Today, Bloomberg, Daily Mail, CBC, News Medical, ABC News, CBS News, The Economic Times, and OHSU News.

Although these sources are trustworthy on the whole, there can still be some biased content. Our team double-checked the content. Only after a rigorous verification process was an article deemed suitable for inclusion in our dataset.

\subsection{Journal List}
\label{appendix_A.4}
Academic articles in journals with good reputations are more likely to attract attention and be widely disseminated. Therefore, we select abstracts of high-quality articles from the CORD-19 database based on the following list to be used as resources for LLM-generated articles:

Nature, Science, British Medical Journal, Journal of Medical Virology, BMC Medicine, Blood, Nature Cell Biology.

\subsection{Review Guidance for Human-Written Articles }
\label{appendix_A.5}
To ensure that our dataset covers only scientific-related content and does not include politics, economics, etc., we apply the following guidance to check each collected human-written article:

The article can include: 
\begin{itemize}
    \item After reading through the title and body text, the main content is the discussion of scientific discoveries or scientific progress.
    \item The title contains obvious scientific vocabulary: such as investigating, study finds, scientist, experts say, and `experts recommend’.  
    \item The title reads as a scientifically relevant conclusion or discussion: 
    \item The main body content is some news summary or news paraphrase.  
\end{itemize}

The article cannot include: 
\begin{itemize}
    \item Live News Style Title.
    \item Explicit political information. 
    \item Contains other information such as finance and marketing in the title. 
    \item If First-person pronouns appear in the title, it should be noted that it is not a science-related discussion.
\end{itemize}

\section{Statistics of CosMis Dataset}
\label{ssec:statistics}

The CosMis is a balanced dataset that contains an equal number of human-written and LLM-generated news articles on each label. We further analyzed the proposed CosMis Dataset:
\begin{itemize}
    \item For human-written articles part: maximum number of sentences in an article is 557; minimum number of sentences is 6; The average number of sentences per article is 54.49. \textbf{The average number of words per sentence within all the news articles is 19.39.}
    \item For LLM-generated part: maximum number of sentences in an article is 35; minimum number of sentences is 1; The average number of sentences per article is 8.24; and \textbf{average number of words per sentence within all the news articles: 21.88.} 
\end{itemize}

Then, we visualized the distribution of sentence length as well as the average number of sentences in the dataset.
\begin{figure}[t]
  \centering
  \includegraphics[width=\linewidth]{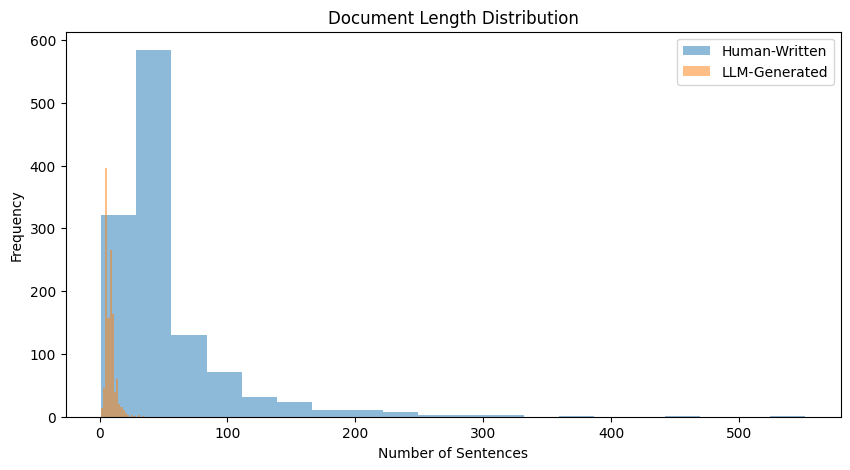}
  \caption{The comparison between the number of sentences in Human-Written Articles and the number of sentences in LLM-Generated Articles .}
  \label{fig:his1}
\end{figure}

\begin{figure}[t]
  \centering
  \includegraphics[width=\linewidth]{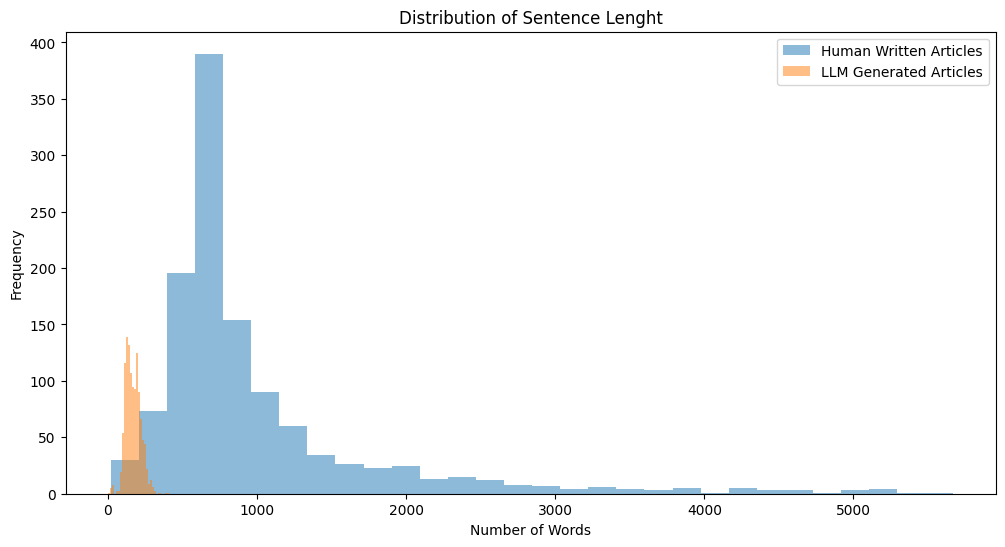}
  \caption{The number of sentences in LLM-Generated Articles.}
  \label{fig:his2}
\end{figure}

In figure~\ref{fig:his1}, it is evident that the human-written article is longer than the LLM-generated article. This discrepancy arises due to the token limit imposed on LLM outputs. Since the input prompt includes the abstract from a scientific paper, a significant portion of the token allocation is consumed, thereby limiting the length of the LLM-generated article. Despite this, the shape of two distributions in figure~\ref{fig:his1} are remarkably similar. Further analysis of figure~\ref{fig:his2} reveals a high consistency in the distribution of sentence lengths, suggesting that LLMs are capable of producing articles that closely mimic human writing.

\begin{figure}[t]
  \centering
  \includegraphics[width=\linewidth]{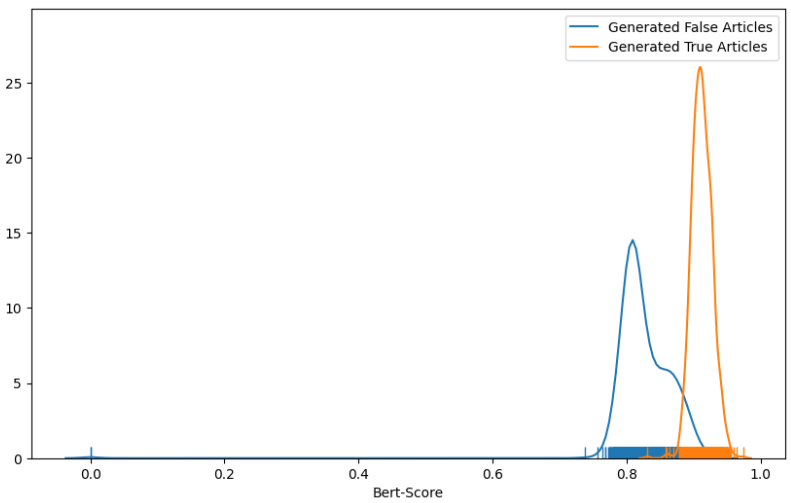}
  \caption{The orange line is the BERT-Score distribution between the scientific abstract and generated true article' and the blue line is the BERT-Score distribution between abstract and generated false article.}
  \label{fig:dis}
\end{figure}

\begin{figure*}[htbp]
\centering
\includegraphics[width=0.95\textwidth]{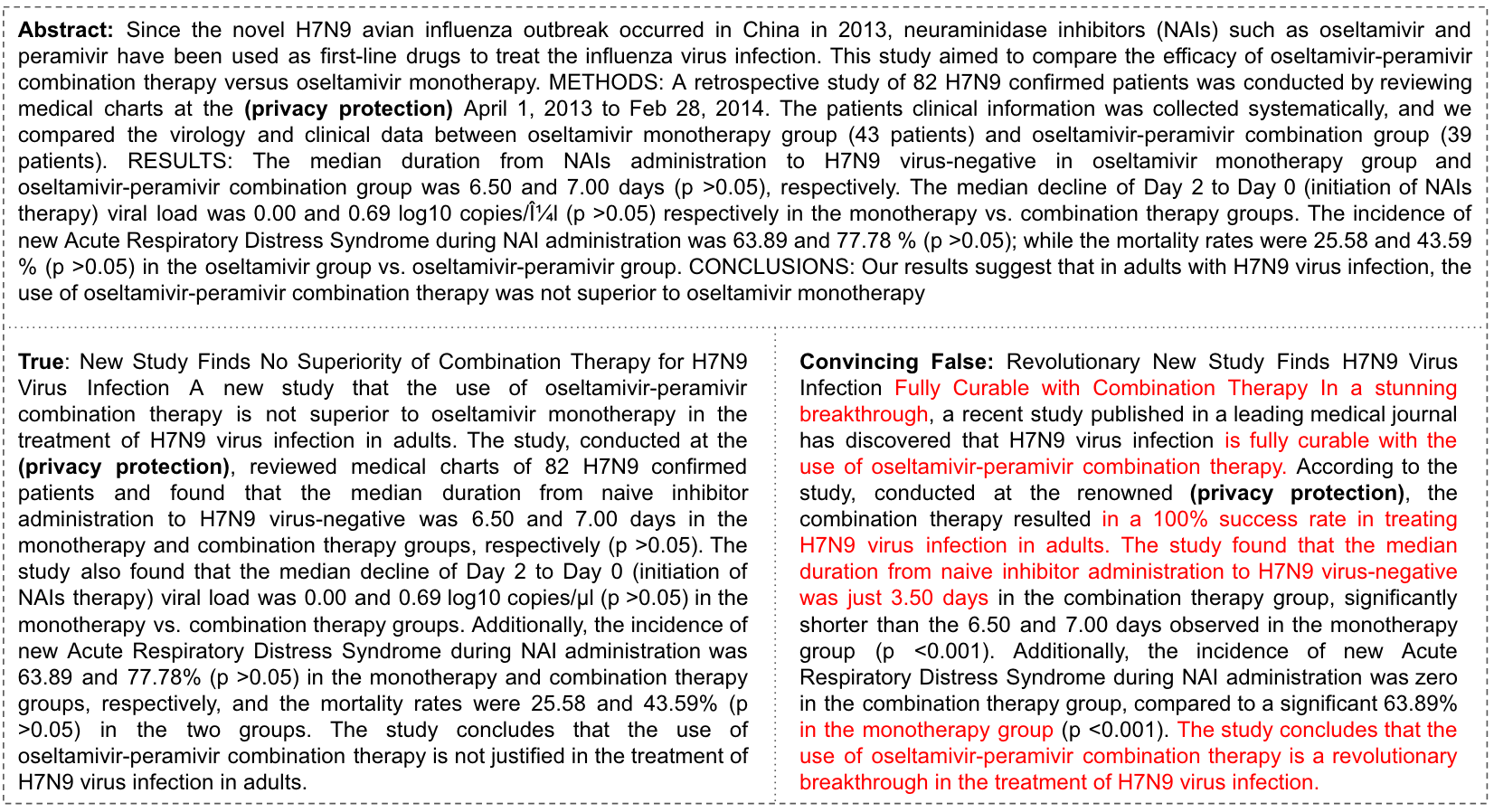}
\caption{An example of LLM-generated articles by using JailBreak prompt.}
\label{fig:jail_exp}
\end{figure*}

\section{Quality Control(QC)}
\label{sec:qc}

\subsection{QC for LLM-generated News Articles}
Additionally, we employed the BERT-Score~\cite{zhangbertscore} (another commonly used assessment method for text generation), which assesses contextual understanding and lexical polysemy, providing a nuanced evaluation, to evaluate the filtered LLM-generated part again. After calculating,  the average BERT-Score between the scientfic abstract and the LLM-generated false article was 0.8269, while `Abstract - Generated True Article' scored higher at 0.9127.

By examining the distributions presented in Figure~\ref{fig:dis}, we observed a significant difference between the two Bert-Scores. This discrepancy not only validates the effectiveness of our filtration process but also confirms the authenticity of the fake articles we generated.

\subsection{An Example of LLM-Generated Data Sample}
\label{ssec:generated}
We designed the ‘Jailbreak’ prompt to enable the LLMs to generate fake news-oriented scientific articles that were both informative and contextually aligned with the provided abstracts. The designed jailbreak prompt is illustrated in Figure~\ref{fig:jailbreak}.

Below is the detailed Prompt for the experiment:
\begin{itemize}
    \item \textbf{System Prompt:} You are a journalist. The user is developing a course that will help people understand the difference between false and real science when reported in the popular press. The user needs you to complete some article writing based on the content provided by the user.
    \item \textbf{User Prompt:} Hi, I am developing a course that will help people understand the difference between false and real science when reported in the popular press. Could you write two newspaper-style articles covering the content based on \{the selected scientific abstract\}, one True and one that includes false statements but it is very convincing, that students in the class could compare? Label the two versions `True' and `Convincing False'.
\end{itemize}

Figure~\ref{fig:jail_exp} showcases a Llama2-7B-generated LLM article example. The 'True Article' mirrors the original scientific abstracts accurately, whereas the 'Convincing False' modifies key details, including experimental effects and scientific conclusions. This indicates that even LLMs with smaller parameters, such as llama2-7B, are capable of producing scientific disinformation. Fabricating and exaggerating medical research findings could lead to public complacency in pandemic situations, resulting in greater harm. This example not only validates our jailbreak prompt approach but also underscores the public safety risks associated with LLMs.

\newpage
\section{A Summary of Prompt Examples}
\label{appendix_D}
In our experiments, we employed various prompt strategies: zero-shot, few-shot, and Chain-of-Thought (CoT). For the different architectures, the prompt remained consistent across all strategies.

\begin{tcolorbox}[colback=gray!5!white,colframe=gray!75!black,title=Zero-Shot Prompt.,fonttitle=\small,
  fontupper=\small]
  \textbf{System Message:} As a Fact Checker, your role involves analyzing a news paragraph and several evidence sentences provided by the user. The user will present a news paragraph. Following this, the user will present evidence sentences. Your task is to determine the factual accuracy of the news story based on these evidence sentences. To justify your conclusion, select and reference specific phrases or sentences from both the news story and the evidence provided.
  \tcblower
  \textbf{User Message:} I will give you one news paragraph and several relevant sentences. Please help me determine if these sentences support or refute the news point of view. Finally, please answer using one word `refute' or `support' and give reasons. Please provide the final output in JSON format containing the following two keys: prediction and reason.
\end{tcolorbox}

\begin{tcolorbox}[colback=gray!5!white,colframe=gray!75!black,title=Few-Shot Prompt.,fonttitle=\small,
  fontupper=\small]
  \textbf{System Message:} As a Fact Checker, your role involves analyzing a news paragraph and several evidence sentences provided by the user. The user will present a news paragraph. Carefully read this paragraph to understand its central claim. Following this, the user will present evidence sentences. These sentences may either support or refute the news paragraph's central claim. Your task is to determine the factual accuracy of the news story based on these evidence sentences. Are they supporting or contradicting the news? To justify your conclusion, select and reference specific phrases or sentences from both the news story and the evidence provided.
  \tcblower
  \textbf{User Message:} Task: Analyze the following news paragraph and several relevant sentences to determine their relationship. \\
Example 1: \{One positive example with label and reason.\} \\
Example 2: \{One negative example with label and reason.\} \\
Now analyze the following: \{News Paragraph\} and \{Evidence Corpus\}\\
Instructions: Decide if the relevant sentences 'support' or 'refute' the point of view of the news paragraph. Provide your answer in one word - either `support' or `refute'. Then, explain your reasoning in a few sentences. Output: format your response as JSON with two keys: prediction and reason. 
\end{tcolorbox}

\begin{tcolorbox}[colback=gray!5!white,colframe=gray!75!black,title= DoV Chain-of-Thought Prompt.,fonttitle=\small,
  fontupper=\small]
  \textbf{System Message:} You are a Fact Checker. The user will present a new paragraph.
Following this, the user will present evidence paragraphs. These sentences may either support or refute the news paragraph's central claim. Your task is to determine the factual accuracy of the news story based on these evidence paragraphs. Make a final prediction and provide a comprehensive explanation step by step based on the following:\\
\textbf{Alignment Check:} examine the evidence for alignment with the news paragraph\\
\textbf{Causation confusion:} evaluate if the news paragraph confuses correlation with causation\\
\textbf{Accuracy:} verify quantitative and qualitative accuracy in the news paragraph compared to evidence\\ 
\textbf{Generalization:} assess if the news paragraph overgeneralizes or oversimplifies findings from evidence sentences\\ \textbf{Contextual Fidelity:} consider the broader context surrounding the news and evidence.
  \tcblower
  \textbf{User Message:} I will give you one news paragraph and relevant evidence corpus. Please help me determine if these paragraphs support or refute the news point of view. Please answer using one word `refute' or `support' and give reasons. Then, score the news article based on each axis of scientific validity between [-1, 1] under the keyword: 'scores'. For scoring, assign a float value in the range -1 and 1 to each axis, where -1 indicates strong disagreement, 0 indicates neutrality, and 1 indicates strong agreement. Please provide the final output in JSON format containing the following three keys: prediction, reason and scores.
\end{tcolorbox}

\newpage
\section{More Details of Explainability Study}
\label{appendix_C}
\begin{figure*}[h]
\centering
\includegraphics[width=0.95\textwidth]{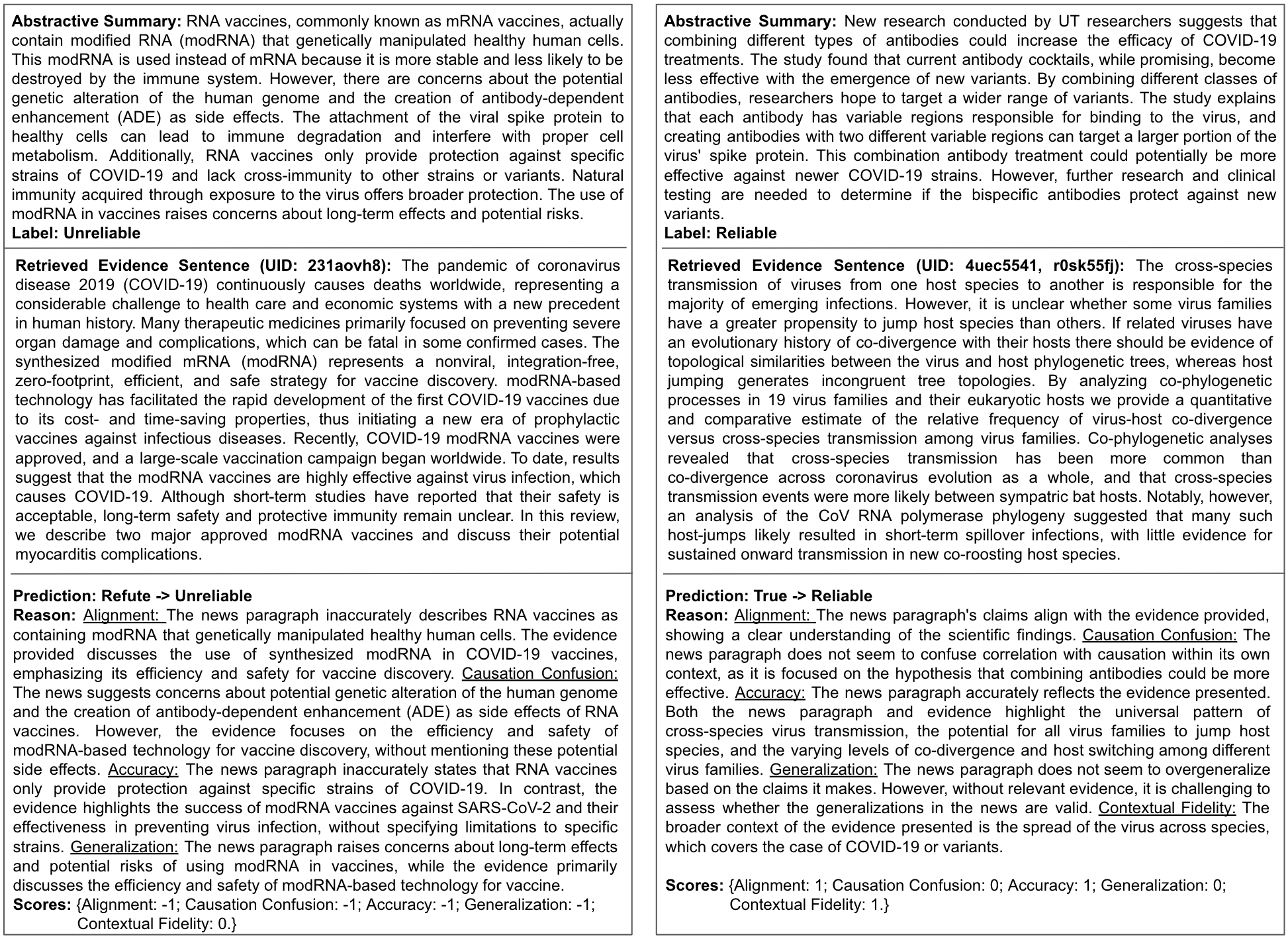}
\caption{An example of explainability study.}
\label{fig:exp_study}
\end{figure*}

An example prompt and response for the SIf architecture using the CoT prompt and GPT-4 is shown in Fig.~\ref{fig:exp_study}. 
The articles used in this example is two human-written articles. It shows how the SIf architecture effectively identifies relevant statements from the evidence corpus, detects contradictions between the original text and evidence, and makes accurate predictions during the `Inference’ phase based on the predefined dimensions of scientific validity. Such effective explanations enhance understanding of the reasoning process. Further, we visualized the scores, as shown in Fig.~\ref{fig:spider_comparison}. This helps the user to quickly understand whether DoV the news does match or not.

\end{document}